\documentclass[conference]{IEEEtran}
\IEEEoverridecommandlockouts
\usepackage{todonotes}
\usepackage{amsmath,amsfonts}

\usepackage{algorithm}
\usepackage{algpseudocode}

\usepackage{array}
\usepackage{textcomp}
\usepackage{stfloats}
\usepackage{url}
\usepackage{verbatim}
\usepackage{graphicx}
\usepackage{cite}
\usepackage{color}
\usepackage{enumitem}
\usepackage{tabularx}
\usepackage{diagbox}
\usepackage{pgfplots, pgfplotstable} 
\pgfplotsset{compat=1.17}
\usepackage{dsfont}
\usepackage{multirow}
\usepackage{lipsum}
\usepackage{booktabs}
\usepackage{ulem}
\usepackage{cite}
\usepackage{mathtools}  
\usepackage{subcaption}

\usepackage{tcolorbox}
\tcbuselibrary{theorems}
\newtcbtheorem{mytheo}{Problem}%
{colback=purple!5,colframe=purple!35!black,fonttitle=\bfseries}{th}

\usepackage{tikz}
\usepackage{ifthen}

\usetikzlibrary{calc}
\usetikzlibrary{math} 
\usetikzlibrary{trees}
\usetikzlibrary{shapes.symbols}
\usepackage{amsthm}
\DeclareMathOperator*{\argmin}{arg\,min}

\usepackage{dsfont}
\usepackage[all]{xy}

\newtheorem{lemma}{Lemma}
\def\BibTeX{{\rm B\kern-.05em{\sc i\kern-.025em b}\kern-.08em
    T\kern-.1667em\lower.7ex\hbox{E}\kern-.125emX}}
\begin{document}

\title{Don't Let Me Down! \\ Offloading Robot VFs Up to the Cloud }

\author{Khasa Gillani\IEEEauthorrefmark{1}\IEEEauthorrefmark{2}, Jorge Martín-Pérez\IEEEauthorrefmark{2}, Milan Groshev\IEEEauthorrefmark{2}, Antonio de la Oliva\IEEEauthorrefmark{2},
Robert Gazda\IEEEauthorrefmark{3}
\thanks{\IEEEauthorrefmark{1}Khasa Gillani is with the NETCOM Lab at IMDEA Networks Institute,
}
\thanks{\IEEEauthorrefmark{2}Khasa Gillani, Jorge Martín-Pérez, Milan Groshev, Antonio de la Oliva are @Departamento de Ingeniería Telemática, Universidad Carlos III de Madrid
}
\thanks{\IEEEauthorrefmark{3}Robert Gazda is with Future Wireless at Interdigital Inc.
}
\thanks{
This work has been partly funded by the Spanish Ministry of Economic Affairs and Digital Transformation and the European Union-NextGenerationEU through the UNICO 5G I+D 6G-EDGEDT and 6G-DATADRIVEN.
}
%
}

\markboth{Journal of \LaTeX\ Class Files,~Vol.~14, No.~8, August~2022}%
{Shell \MakeLowercase{\textit{et al.}}: Application Aware Networking for Robotic Applications}


\maketitle

\begin{abstract}

Recent trends in robotic services propose offloading robot
functionalities to the Edge to meet the strict latency requirements of
networked robotics.
However, the Edge is typically an expensive resource and
sometimes the Cloud is also an option, thus, decreasing
the cost.
Following this idea, we
propose Don't Let Me Down! (DLMD),
an algorithm that promotes offloading robot functions to the Cloud
when possible to minimize
the consumption of Edge resources.
Additionally, DLMD takes the appropriate
migration, traffic steering, and radio
handover decisions to meet robotic service
requirements as strict latency constraints.
In the paper we formulate the
optimization problem that
DLMD aims to solve, compare DLMD performance
against the state of the art, and
perform stress tests to assess DLMD
performance in small \& large networks.
Results show that DLMD ($i$) always
finds solutions in less than 30ms;
($ii$) is optimal in a local warehousing
use case; and ($iii$)
consumes only 5\% of the Edge resources
upon network stress.

\end{abstract}


\section{Introduction}
\IEEEPARstart{N}{etworked} robotic services are being adopted to enhance operational automation and performance in some uses cases,
e.g., assembly robots in Industry 4.0,
or remotely controlled robots.
However, in such use cases, strict latency requirements~\cite{3GPProbots} are difficult to meet
upon network congestion or large latencies
towards the servers hosting the networked
robotic services.

To overcome such limitation,
recent works~\cite{delgado2022oros, chen2018qos} propose to split the networked robotics
functionality into Virtual Functions (VFs), and offload them to servers with more computational resources.
Offloading VFs of a robotic service implies solving the VF embedding problem. A plethora of works in the literature have tackled the problem during the last years using artificial intelligence~\cite{nfvdeep} and bin packing-alike heuristics~\cite{joint}. The solutions guarantee that resources – e.g. bandwidth and CPUs – are not exhausted, and typically minimize the latency
of the embedded service. Recent works have adapted the VF embedding problem to robotic services -- see~\cite{delgado2022oros,chen2018qos,delayandreliability} -- but failed to consider either the latency~\cite{delgado2022oros};
radio signal quality~\cite{delayandreliability,chen2018qos}; or robot
mobility~\cite{chen2018qos}. Consequently, the robotic service may not fulfill strict service latency requirements, suffer from low bitrates, or service disruption -- due bad radio connectivity/coverage.

\begin{figure}[t]%
    \subfloat[\label{fig:algosc} ]{{\includegraphics[width=0.35\columnwidth]{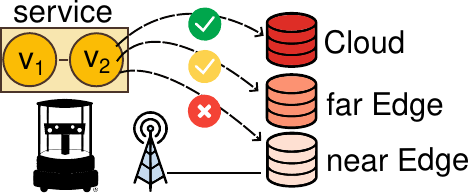} }}~
    \subfloat[\label{fig:avoid} ]{{\includegraphics[width=0.24\columnwidth]{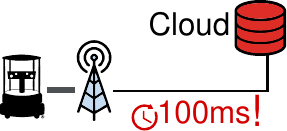} }}~
    \subfloat[\label{fig:avoid-coverage} ]{{\includegraphics[width=0.24\columnwidth]{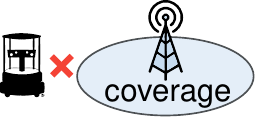} }}

    \caption{Don't Let Me Down! fosters
    offloading the robot service VFs up to
    the cloud (a), yet preventing the cloud
    large latencies (b)
    and running out of coverage (c).}
    \label{fig:dlmd}%
\end{figure}

To that end, in this paper we propose Don't Let Me Down! (DLMD), an algorithm that fosters offloading robotic VFs to the Cloud to minimize the Edge usage while considering ($i$) the robotic service latency constraints;
($ii$) the wireless connectivity; and
($iii$) robot mobility. DLMD fosters offloading the VFs to the Cloud as long as the latency constraints are fulfilled, and takes VF migration and radio handover decisions to prevent large latencies and out of coverage situations -- see~Fig.~\ref{fig:dlmd}.

\section{Don't let me Down! problem formulation}
\label{sec:optimization}

In this section we formulate the VF embedding problem that DLMD solves to
take adequate offloading, migration and radio handover decisions
for robotic services.
We consider a hardware graph $G$ whose
vertices $V(G)$ correspond to switches and
servers. Specifically, we consider the
three tiers of servers illustrated in
Fig.~\ref{fig:algosc}: Cloud, far Edge, and
near Edge; each with decreasing latency
towards the robot, respectively.

The goal of the VF embedding problem
is to offload robot VFs 
minimizing the resource
consumption at the Edge, and satisfying
the robotic service constraints.
In the following
Sections~\ref{subsec:computational-constraints}
to \ref{subsec:radio-constraints} we specify the robotic service constraints, and in
Section~\ref{subsec:aan-complexity} we
formulate the associated
VF embedding problem statement, and
prove its NP-hard complexity.

\subsection{Robot computational constraints}
\label{subsec:computational-constraints}
The VF embedding problem must ensure that the robot VFs do not exhaust the computational resources $C(n)$ of each server $n\in V(G)$:
\begin{equation}
    \sum_{v \in a(n)} C(v) \leq \ C(n) , 
 \quad \forall n \in V(G) 
    \label{eq:computational-constraint}
\end{equation}
That is, the computational requirements of all VFs $v$ assigned to a computing node $a(n)=\{v_1,v_2,\ldots\}$ must be lower than its available computational resources $C(n)$.
On top, it must ensure that all VFs $V(S_i)=\{v_1,v_2,\ldots\}$ of a robotic service $S_i$ are offloaded at some computing node $n$:
\begin{equation}
    \sum_{n \in V(G)} P(v,n) \geq 1 , \quad \forall {S_i}\in\mathcal{S}, v \in V({S_i})
    \label{eq:at-least-once}
\end{equation}
with $P(v,n)=1$ if VF $v$ is offloaded at the computing node $v\in a(n)$, and $0$ otherwise.
Note that we do not prevent a VF to be ``replicated'' -- or offloaded in multiple nodes --
since load-balancing may be required by some robotic services coordinating multiple robots.

\subsection{Network constraints}
\label{subsec:network-constraints}
For VFs are offloaded to the Cloud or Edge
servers, the robotic service traffic has to
be steered across the network.
The steering cannot exhaust the bandwidth $\lambda(n_1,n_2)$ over the
hardware links -- i.e., edges $E(G)$ of the hardware graph:
\begin{equation}
    \sum_{\mathclap{(v_1,v_2) \in a(n_1,n_2)}} \lambda (v_1,v_2)\leq (1-\delta(n_1,n_2)) \lambda(n_1,n_2),\\\forall (n_1,n_2) \in E(G)\\
    \label{eq:bandwidth-constraint}
\end{equation}
with $\delta(n_1,n_2)\in[0,1]$ the packet drop rate 
present at the link
$(n_1,n_2)\in E(G)$.
Specifically, all virtual links (VL) of
the service $S_i\in\mathcal{S}$
should not use all the link bandwidth $\lambda(n_1,n_2)$.
In~\eqref{eq:bandwidth-constraint},
$a(n_1,n_2)=\{(v_1,v_2),(v_2,v_3),\ldots\}$
denotes the VLs
assigned/traversing the link $(n_1,n_2)$.

Also -- inline with~\eqref{eq:at-least-once} -- all the VLs $E(S_i)$
traffic must be processed at the server(s) where the VFs are offloaded, i.e.:
\begin{equation}
    \displaystyle \sum_{\mathclap{(n_1,n_2) \in E(G)}} P(v_1,v_2,n_1,n_2)\geq 1, \quad \forall {S_i}\in\mathcal{S}, (v_1,v_2) \in E({S_i})
    \label{eq:vl-at-least-once}
\end{equation}
with $P(v_1,v_2,n_1,n_2)=1$ if the VL $(v_1,v_2)$ is assigned to a link $(v_1,v_2)\in a(n_1,n_2)$, and
$P(v_1,v_2,n_1,n_2)=0$ otherwise.
Note that constraint~\eqref{eq:vl-at-least-once} also
ensures that every VL of a robotic service $E(S_i)$ traverses at least one physical link $(n_1,n_2)$.

Upon the VF offloading, the VLs steering must satisfy the
flow constraint~\cite{approx-algs}. That is, every switch $w_i$ and Point
of Access (PoA) $R_i$
ingresses and egresses the same amount of traffic:
\begin{equation}
    \qquad \sum_{\mathclap{\substack{(v_1,v_2)\in a(n_1,n):\\(n_1,n)\in E(G)}}}\lambda(v_1,v_2) = \sum_{\mathclap{\substack{(v_1,v_2)\in a(n,n_2):\\(n,n_2)\in E(G)}}}\lambda(v_1,v_2), \quad \forall n\in \{w_i\} \cup \{R_i\}
    \label{eq:flow-constraint}
\end{equation}


It is also important that every
server $\forall n_2\in\{s_i\}_i$ processing VF
$v_2$ should receive the corresponding VL traffic:
\begin{equation}
\begin{split}
    \sum_{\mathclap{(n_1,n_2) \in E(G)}} P(v_1,v_2,n_1,n_2)= P(v_2,n_2) \quad\forall (v_1,v_2) \in V({S_i})
    \end{split}
    \label{eq:steer-to-vf}
\end{equation}
otherwise, a solution of the problem may mistakenly steer the traffic to a server
without VF $v_2$ offloaded there.

\subsection{Robot delay constraints}
\label{subsec:delay-constraints}
To meet the latency constraints of a robotic service,
it is necessary to consider both
the propagation and processing delay perceived by the robot when consuming the service.

The network delay experienced by the robotic service $S_i$ is the sum of the delay of links $d(n_1,n_2)$ traversed by the VLs $(v_1,v_2)$,
and the queuing delay $\psi(n_1,n_2)$ that
may be present in network:
\begin{equation}
    d_{net} (S_i) =\sum_{(v_1, v_2)\in E(S_i)} \sum_{\mathrlap{\substack{(n_1,n_2)\in E(G)\colon\\ (v_1,v_2)\in a(n_1,n_2)}}} d(n_1,n_2)+\psi (n_1,n_2),\forall S_i\in\mathcal{S}
    \label{eq:network-delay}
\end{equation}
To compute the processing delay we resort to the
\mbox{M/G/1-PS} expression for the average delay as it's a common practice in the existing literature\cite{RCohen15,jemaa2016qos,oljira2017model}. Therefore, we obtain the processing delay of a VF $v$ as:
\begin{equation}
d_{pro} (v) =  \sum_{\mathclap{(v_1,v)\in E(S_i)}}\quad\frac 1{C(v)\mu-\lambda (v_1,v)},\quad \forall S_i\in\mathcal{S}, v\in V(S_i)
\end{equation}
where $\mu$ is the processing rate of a CPU.
That is, we have an \mbox{M/G/1-PS}
system with an aggregate processing rate $C(v)\mu$
and an arrival rate $\lambda(v_1,v)$, i.e., the incoming traffic to the VF $v$.
Hence, any offloading solution
must ensure that
the network and processing delay remain below the
requirement of the robotic service $D(S_i)$:
\begin{equation}
    d_{net}(S_i) + \sum_{v\in V(S_i)}d_{pro}(v) \leq D(S_i), \quad \forall S_i\in\mathcal{S}
    \label{eq:latency-constraint}
\end{equation}

\subsection{Robot radio constraints}
\label{subsec:radio-constraints}
Since robotic services leverage wireless
technologies to connect with the offloaded VFs,
the offloading must prevent using radio links that cannot meet
the robotic service requirements.

The first constraint to impose is that
a VL $(v_1,v_2)$ cannot traverse the link connecting the robot with the PoA $(r_i,R_i)$ unless the robot
wireless interface is attached to the PoA:
\begin{equation}
    P(v_1,v_2,r_i,R_i) \leq \phi(r_i,R_i), \quad \forall (v_1,v_2), r_i,R_i
    \label{eq:steer-if-attached}
\end{equation}
with $\phi(r_i,R_i)=1$ if the offloading solution
tells the robot $r_i$ to attach to the PoA $R_i$,
and zero otherwise. Note that $\phi(r_i,R_i)$ also represents
the robot handover across PoAs as it moves.
Any offloading solution must ensure
that the robot $r_i$ network interface is attached to one PoA $R_i$ to have connectivity:
\begin{equation}
    \sum_{R_i} \phi(r_i,R_i) = 1, \quad \forall r_i\in\{r_i\},i
    \label{eq:attach-to-one}
\end{equation}
otherwise, any robotic service $S_i$ will not have
connection to the remote server where the VF(s) are offloaded.

Since the wireless connectivity suffers from background noise~$N$ and heavily depends on the signal strength~$\sigma_{R_i}(r_i)$, it is necessary to account for the effective bandwidth capacity over the wireless link from the robot to the
PoA $(r_i,R_i)$. Inline with recent works as~\cite{cellos}, we model the wireless transmission
capacity as:
\begin{equation}
    \mathrm{T}(r_i,R_i) = (1-\delta(r_i,R_i))\lambda(r_i,R_i)\log_2\left(1+\frac{ \sigma_{R_i}(r_i)}{N}\right)
    \label{eq:channel-capacity}
\end{equation}
with $(1-\delta(r_i,R_i))\lambda(r_i,R_i)$ being
the perfect conditions bandwidth over the wireless
link $(r_i,R_i)$, considering the packet loss $\delta(r_i,R_i)$; and
the $\log_2(1+\sigma_{R_i}(r_i)/N)$ being the attenuation term under the presence of noise given a certain signal strength. Note that we assume $N$
is additive Gaussian white noise.

Knowing the bandwidth constraint discussed
in~\eqref{eq:bandwidth-constraint}, if one accounts
for the wireless transmission capacity $T(r_i,R_i)$
of the link in between the robot $r_i$ and the
PoA $R_i$, the following must hold:
\begin{equation}
    \sum_{\mathclap{(v_1,v_2)\in a(r_i,R_i)}} \lambda(v_1,v_2) \leq \mathrm{T}(r_i, R_i), \quad \forall (v_1,v_2)\in a(r_i,R_i)
    \label{eq:wireless-bandwidth-capacity}
\end{equation}
so all VLs $(v_1,v_2)$ traversing the robot-to-PoA
wireless connection do not exceed the transmission
capacity. 

\begin{figure}[t]
        \centering
        \begin{tikzpicture}[scale=.8]
    \tikzmath{\xshift=.4; \yshift=1;};

    \tikzmath{\xrobot=6; \yrobot=-2*\yshift;}
    \node[fill=white,draw=black,rectangle] (robot) at (\xrobot, \yrobot) {$r_1$};
    \node[fill=white,draw=black,rectangle] (poa) at ($(robot)+(1.5,0)$) {$R_1$};
    \draw[fill=white,draw=black] (robot) -- (poa);
    \node[fill=white,draw=black,rectangle] (s1) at ($(poa)+(2.2,1)$) {$s_1$};
    \node[fill=white,draw=black,rectangle] (sinf) at ($(poa)+(2.2,-1)$) {$s_\infty$};
    \node (sdots) at ($(poa)+(1.5,0)$) {$\vdots$};
    \draw[draw=black] (poa) -- (s1);
    \draw[draw=black] (poa) -- (sinf);

    \foreach \j in {4,...,1}{
        \tikzmath{\x=\j*\xshift; \y = -1 * \j * \yshift; }
        \node (vr) at (\x, \y) {};
        \node (v2) at ($(vr) + (2,0)$) {};

        \ifthenelse{\j=4}{
            \draw[draw opacity=0] (vr) -- (v2) node[pos=0.5]{$\ldots$};
        }{
            \draw[draw=black,fill=white] ($(vr) - (1.2,.8)$) rectangle ++(4.7,1.7);
            \draw (vr) -- (v2) node[pos=-0.3,circle,draw]{$v_{r,\j}$} node[pos=1.3,circle,draw]{$v_{2,\j}$};
            \node at ($(vr) + (3.2,.6)$) {$S_\j$};

            \draw[->,dotted,thick] ($(vr) - (.3,.4)$) --  ($(vr)-(.3,.4)$) -|  node[pos=.85,below,anchor=north west]{\ifthenelse{\j=3}{\small $P(v_{r,\cdot},r_1)=1$}{}} (robot);

            \ifthenelse{\j=1}{
                \draw[dashed,->] ($(v2) + (.4,.4)$) -- ($(v2)+(.4,1.1)$) --  node[midway,above] {$P(v_{2,1},s_1)=?$} ($(s1)+(0,1.1)$) -- (s1);
            }{}
            \ifthenelse{\j=2}{
                \draw[dashed,->] ($(v2) + (.8,0)$) -- ($(s1)-(4,0)$)  -- node[midway,above] {$P(v_{2,2},s_1)=?$} (s1);
            }{}
            \ifthenelse{\j=3}{
                \draw[->,dashed] ($(v2) + (.8,0)$) -- ($(v2)-(-.8,1.2)$) -- node[pos=.7,above] {$P(v_{2,3},s_\infty)=?$} ($(sinf)-(0,1.2)$) -- (sinf);
            }{}

        }
    }

\end{tikzpicture}
        \vspace{-1.5em}
        \caption{VF embedding in a network with one robot
        $r_1$, one PoA $R_1$, and an infinite pool
        of servers $\{s_i\}_i^\infty$ to decide which
        server we offload the second VFs
        $P(v_{2,\cdot},s_i)$ is the bin-packing problem, thus,
        it is NP-hard.}
        \label{fig:np-hard}
    \end{figure}
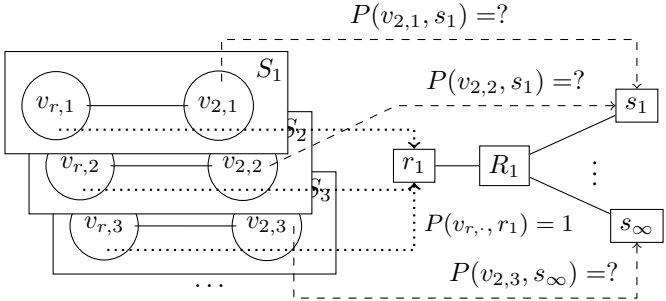
\subsection{Problem statement and complexity}
\label{subsec:aan-complexity}
With the prior constraints we formulate the
optimization problem that captures the
associated VF embedding problem.
The goal is to minimize the used Edge resources,
so the substrate network is shared by multiple robotic service.


\begin{mytheo}[label={problem}]{VF embedding for robotic services}{theoexample}
    Given the
    computational~\mbox{\eqref{eq:computational-constraint}-\eqref{eq:at-least-once}},
    network~\mbox{\eqref{eq:bandwidth-constraint}-\eqref{eq:steer-to-vf}},
    delay~\mbox{\eqref{eq:latency-constraint}},
    and radio constraints~\mbox{%
        \eqref{eq:steer-if-attached},
        \eqref{eq:attach-to-one},
        \eqref{eq:wireless-bandwidth-capacity}
    }
    of a robotic service~$\mathcal{S}$;
    minimize the Edge resource usage 
    with adequate VF offloading $P(v,n)$,
    routing $P(v_1,v_2,n_1,n_2)$, and
    attachment $\phi(r_i,R_i)$ decisions.
    \begin{align}
        \min_{P(\cdot), \phi(\cdot)} & \sum_{n \in V(G)} \kappa_{n} |a(n)| 
        \nonumber\\
        s.t.: &\quad \eqref{eq:computational-constraint}-\eqref{eq:steer-to-vf},\ \eqref{eq:latency-constraint}-\eqref{eq:attach-to-one},\ \eqref{eq:wireless-bandwidth-capacity}\nonumber
    \end{align}
    with $\kappa_n\in\mathbb{N}$ being the server cost, which
    takes higher values if the server $n$ is closer to the Edge.
\end{mytheo}

We solve Problem~\ref{problem} iteratively as the robot moves.
Hence, changes in $\phi(\cdot)$ , and $P(\cdot)$ decision
variables represent handovers and VF migrations respectively.
However, finding the optimal solution of Problem~\ref{problem} is not straight-forward.
Inline with the complexity of existing optimization
problems~\cite{sang2017provably,approx-algs},
in the following Lemma~\ref{lemma} we prove that
it is NP-hard.

\begin{lemma}
    Solving Problem~\ref{problem} is NP-hard.    \label{lemma}
\end{lemma}

\begin{proof}
    We proof that our proposed problem is NP-hard showing
    that an instance of Problem~\ref{problem} is
    equivalent to the bin-packing problem ~\cite{complexity}.
    Let's consider a set of ``ideal'' robotic services~$\mathcal{S}'$
    without any delay $D(S_i)=\infty,\ \forall S_i\in\mathcal{S}'$ and
    bandwidth $\lambda(v_1,v_2)=0$ requirements. On top, these ``ideal'' robotic
    services consist of
    just two VFs $v_{r,i},v_{2,i}=V(S_i),\ \forall S_i\in\mathcal{S}'$,
    a single VL $(v_{r,1},v_{2,i})$
    and the former VF has to run in the robot
    $P(v_{r,i},r_1)=1,\ \forall S_i\in\mathcal{S}'$
    -- see~(Fig.~\ref{fig:np-hard}). Hence
    constraints \eqref{eq:at-least-once} and \eqref{eq:vl-at-least-once}
    are strict equalities.
    
    Now lets assume that all these ``ideal'' robot
    services have to be deployed in a hardware graph
    consisting of one robot $r_1$ with infinite
    computational capacity $C(r_1)=\infty$, a single
    PoA $R_1$ that covers all the geographical area
    with adequate signal strength towards the robot,
    so $\sigma_{R_1}(r_1)>N$ holds, and an infinite
    number of servers $\{s_i\}_i^\infty$ with
    finite capacity $C(s_i)<K\in\mathbb{N}^+$
    and same cost $\kappa_{n_1}=\kappa_{n_2},\forall n_1,n_2\in V(G)$.
    All the servers have at one hop distance the PoA
    $R_1$, as shown in~Fig.~\ref{fig:np-hard}.

    Hence, in the resulting scenario -- depicted
    in~Fig.~\ref{fig:np-hard} -- the VF embedding consists in deciding where to deploy
    the second VF, i.e., $P(v_{2,i},s_i),\ \forall S_i\in\mathcal{S}'$.
    For the VL routing decision will be to traverse
    the single link connecting the PoA with the
    server, i.e.,
    $P(v_{2,i},s)=1\implies P(v_{r,i},v_{2,i},R_1,s)=1,\ \forall S_i\in\mathcal{S}'$
    with $s\in\{s_i\}_i^\infty$.
    As a result, the considered ``ideal'' robot
    service leads to the following instance of
    Problem~\ref{problem}:
    \begin{align}
        \min_{P(v_{2,i},s)} & \sum_{s\in \{s_i\}_1^\infty}
        |a(s)|\label{eq:pro2-obj}\\
        s.t.: 
        &\sum_{v_{2,i} \in a(s)} C(v_{2,i})\leq C(s),\quad \forall s\in \{s_i\}_1^\infty\label{eq:pro2-capacity}\\
        & \sum_{s\in\{s_i\}_1^\infty} P(v_{2,i},s) = 1,\quad \forall S_i \in\mathcal{S}'\label{eq:pro2-deploy}
    \end{align}
    where the bandwidth constraints
    \eqref{eq:bandwidth-constraint},
    \eqref{eq:flow-constraint} and~\eqref{eq:wireless-bandwidth-capacity}
    do not apply given that
    $\lambda(v_{r,i},v_{2,i})=0,\ \forall S_i\in\mathcal{S}'$.
    Note that the radio attachment decision is given,
    hence, \eqref{eq:steer-if-attached} and
    \eqref{eq:vl-at-least-once} are satisfied;
    and the delay constraint~\eqref{eq:latency-constraint}
    is always satisfied for the ``ideal'' robotic service.
    $D(S_i)=\infty,\ \forall S_i\in\mathcal{S}'$.
    
    Overall, \eqref{eq:pro2-obj}-\eqref{eq:pro2-deploy}
    is an instance of Problem~\ref{problem} that
    is equivalent to the bin-packing
    problem with servers $\{s_i\}_1^\infty$ as bins
    with size $C(s)$, and the second VFs $v_{2,i}$
    as items with size $C(v_{2,i})$.
    Therefore, Problem~\ref{problem} is NP-hard.
\end{proof}

\section{Don't Let Me Down!: the algorithm}
\label{sec:context-fp}
DLMD solves Problem~\ref{problem} by offloading the VFs up to the Cloud if its delay is not too large. First, DLMD selects which are the PoAs covering the robot,
and offering enough wireless capacity for the first VL $(v_1,v_2)$
-- see line~\ref{alg:prune} in Algorithm~\ref{alg:cap}.
Second, DLMD decides to which server it should offload every VF.
To do that, DLMD keeps a metric $\tau$ representing the trade-off between the
server cost $\kappa_n$,
the free bandwidth of links to reach the server $\tfrac{1}{\lambda(n_1,n_2)}$,
and the delay of such links $d(n_1,n_2)$. Specifically, $\tau$ is derived
using Dijkstra with
weight $1/\lambda(\cdot) + d(\cdot)$ towards the candidate servers where VF are
offloaded -- see line~\ref{alg:dijkstra} of Algorithm~\ref{alg:cap}.

Third, DLMD offloads the VF to the server with best $\tau$ metric, and steers the
traffic over the path found by the $\tfrac{1}{\lambda}+d$-weighted Dijkstra.
Lastly, DLMD selects the PoA with highest bandwidth available and small delay
to the robot -- see line~\ref{alg:select-poa} of Algoritm~\ref{alg:cap}.

\algrenewcommand\algorithmicrequire{\textbf{Input:}}
\algrenewcommand\algorithmicensure{\textbf{Output:}}

\begin{algorithm}[H]
\caption{Don't Let Me Down! (DLMD)}\label{alg:cap}
\begin{algorithmic}[1]

\Require $G,S_i,r_i, \{R_i\}_i, \{\kappa_n\}_{n\in V(G)}$
\Ensure $\{\phi(r,R_i)\}_i,\ \{P(v_i,n_i)\}_i,\ \{P(v_i,v_j,n_i,n_j)\}_{i,j}$

\State $\{\hat{R}_i\} = \{R_i: \lambda(v_1,v_2) > T\}_i$ \label{alg:prune}

\For{$v\in V(S_i)$}
    \State $\tau_{\min} = \infty$
    \For{$n\in V(G)$}
    \State $\tau, p = \kappa_n + $ Dijkstra($n,n_{-1}$, weight=$\tfrac{1}{\lambda}+d$)\label{alg:dijkstra}
        \If{$\tau < \tau_{\min}$}
            \State $\tau_{\min}=\tau$
            \State $P(v,n)=1$
            \State $\{P(v_{-1}, v, n_1, n_2)=1\}_{(n_1,n_2)\in p}$
        \EndIf
    \EndFor

    \State $\phi\left(r,\argmin_{\hat{R}_i} \tfrac{1}{\lambda(r_i,\hat{R}_i)}+d(r_i,\hat{R}_i))\right) = 1 $\label{alg:select-poa}
\EndFor

\end{algorithmic}
\end{algorithm}

Thanks to its trade-off metric $\tau$, DLMD will always try to offload the VFs
to the Cloud unless there are Edge servers with significantly smaller latency and leaves as much free resources in the Edge as possible
-- thus, minimizing Problem~\ref{problem} objective --
and steer the traffic over non-congested links. DLMD is invoked iteratively to update the offloading $P(\cdot)$ and
attachment/handover $\phi(\cdot)$ decisions as robot moves.

\section{Experimental Results}

\newcommand{\delaytab}{
    \begin{tabular}{c l l l }
        \toprule
         {\bf PoA} & {\bf Cloud} & {\bf  fEdge} & {\bf\ nEdge} \\ 
        \midrule
        R\textsubscript{1}, R\textsubscript{3} & 9~ms & 4~ms & 3~ms\\
       R\textsubscript{2}, R\textsubscript{4} & 18~ms & 8~ms & 9~ms\\
       R\textsubscript{5}, R\textsubscript{6} & 27~ms & 12~ms & 9~ms\\
       \bottomrule
    \end{tabular}
}

\begin{figure}[t]
    \centering
    \subfloat[\label{fig:corridor} ]{{\includegraphics[width=0.5\columnwidth]{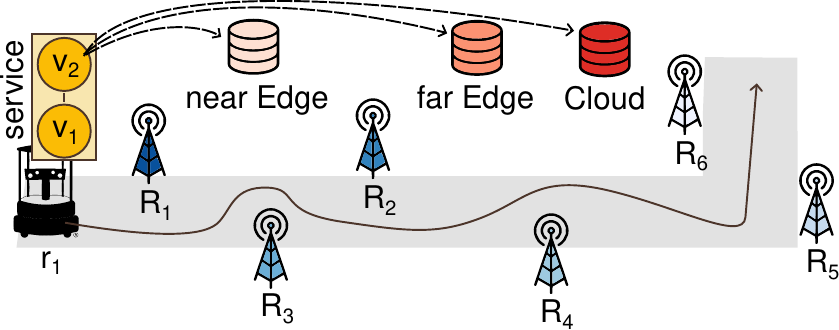} }}~
    \subfloat[\label{tab:delays}]{\vspace{1em}\resizebox{.45\columnwidth}{!}{\delaytab}}
    
    \subfloat[\label{fig:oros} ]{{\includegraphics[trim=27 5 26 14,clip,width=0.31\columnwidth]{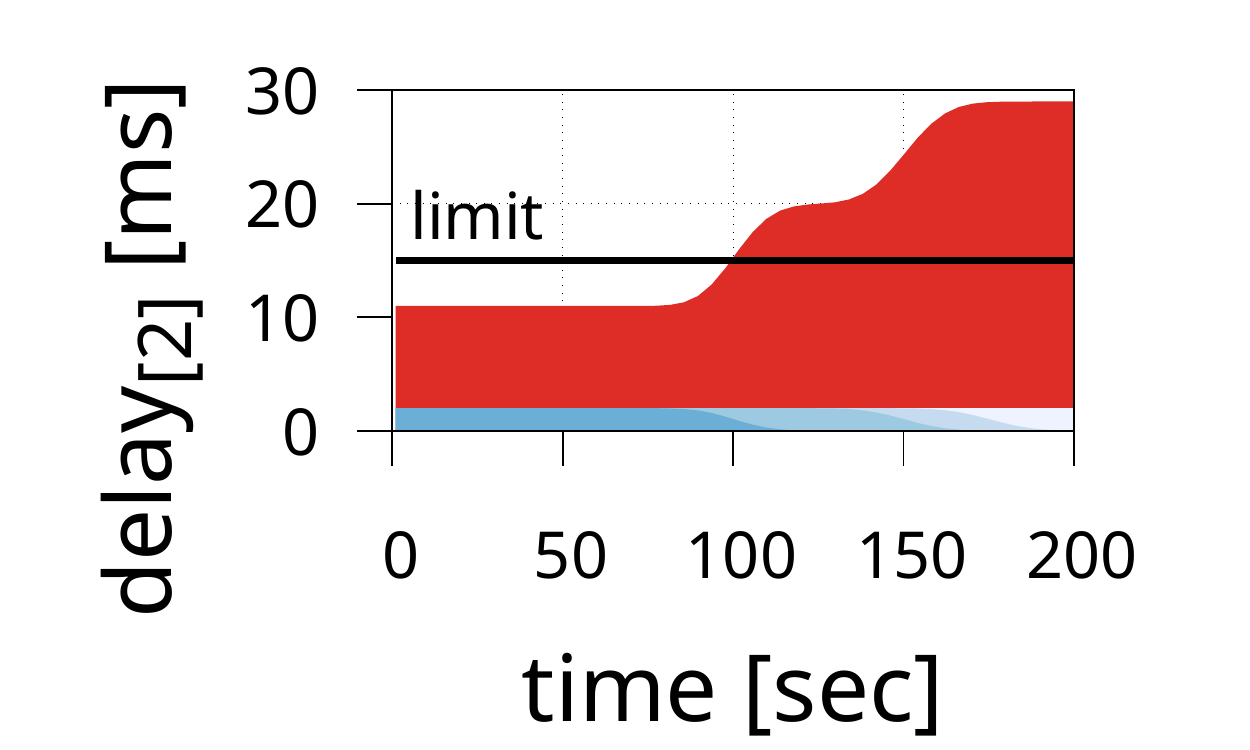} }}~
    \subfloat[\label{fig:tmc} ]{{\includegraphics[trim=27 5 26 14,clip,width=0.31\columnwidth]{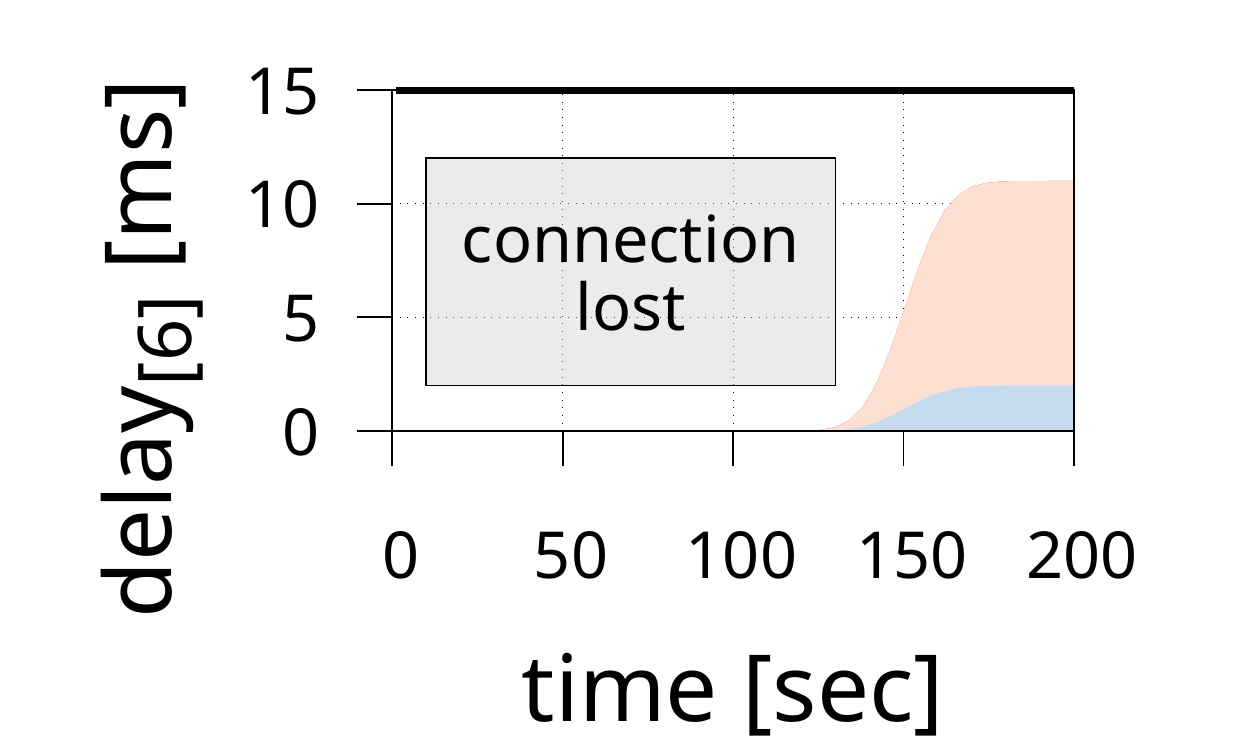} }}~
    \subfloat[\label{fig:dlmd-results} ]{{\includegraphics[trim=27 5 26 14,clip,width=0.31\columnwidth]{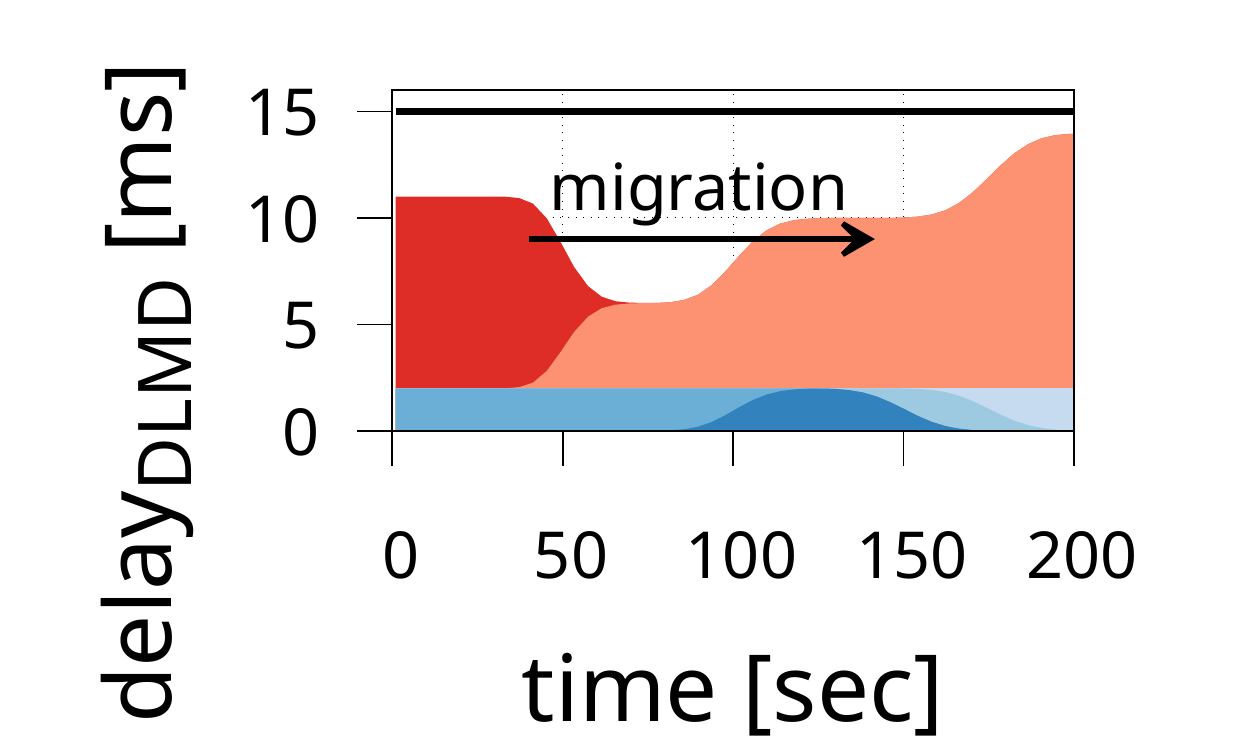} }}~
    
    \caption{Delay experienced by the robot
    as it drives in a warehousing 
    scenario (a). (b)-(d) illustrate
    the delay contribution of PoAs (blue fill)
    and servers (red fill) using
    \cite{delgado2022oros},
    \cite{delayandreliability} and DLMD.}
    \label{fig:exp-seup}
\end{figure}

In this section we assess DLMD performance in two scenarios:
$(i)$ a small warehousing scenario taking VF offloading, migration, and handover decisions; and $(ii)$ a stress test in small and large graphs with real-world PoA locations.

\subsection{Small warehousing scenario}
We consider a factory floor with a warehousing robot service $S_1$
that offloads its remote driving
VF $v_2$ to a near/far Edge or Cloud server
-- with light-, mid- and intense-red in Fig.~\ref{fig:corridor}; respectively.
As the robot moves, 
DLMD must decide to which PoA $R_1,\ldots,R_6$ (blue color) it should attach, and make the
corresponding offloading and migration decisions for the remote driving VF $v_2$.
Note that the robot drivers VF is denoted as $v_1$
and they reside in the robot in the considered
experiment.

The initial offloading, migration and handover
decisions must be such that the service latency
remains below the $D(S_1)=15$~ms requirement of the
considered warehousing service.
Fig.~\ref{tab:delays} reports the different delays
that each PoA has towards the servers. Figs.~\ref{fig:oros}-\ref{fig:dlmd-results} show
the delay experienced by the warehousing robot
during its trajectory
when we use \cite{delgado2022oros},
\cite{delayandreliability} and DLMD; respectively.

Results show that \cite{delgado2022oros}
violates the 15~ms limit when the robot
is half-way to the end of its trajectory
(100~sec.) because it connects to
PoAs $R_5, R_6$ with high latency towards the
Cloud server where $v_2$ is offloaded
-- see~Fig.~\ref{fig:oros}. When using \cite{delayandreliability} in experiments,
the robot lost connectivity during the first
125~sec. -- see~Fig.~\ref{fig:tmc}. The reason is that
\cite{delayandreliability} neglected the bad SNR
of the PoAs, and resulted into trying to steer
traffic over wireless links without enough capacity
due to the bad signal conditions.

The aforementioned problems were not experienced
by DLMD -- see~Fig.~\ref{fig:dlmd-results} --, for
it instructs the robot to attach to PoAs with
adequate radio conditions, and migrated the remote
driving VF $v_2$ to the far Edge when the robot
attached to PoAs with high latency towards the Cloud
to meet the 15~ms limit.

\begin{figure}[t]
    \centering
    \subfloat[\label{fig:compare-snr}   ]{{\includegraphics[trim=13 21 18 10, clip,width=.42\columnwidth]{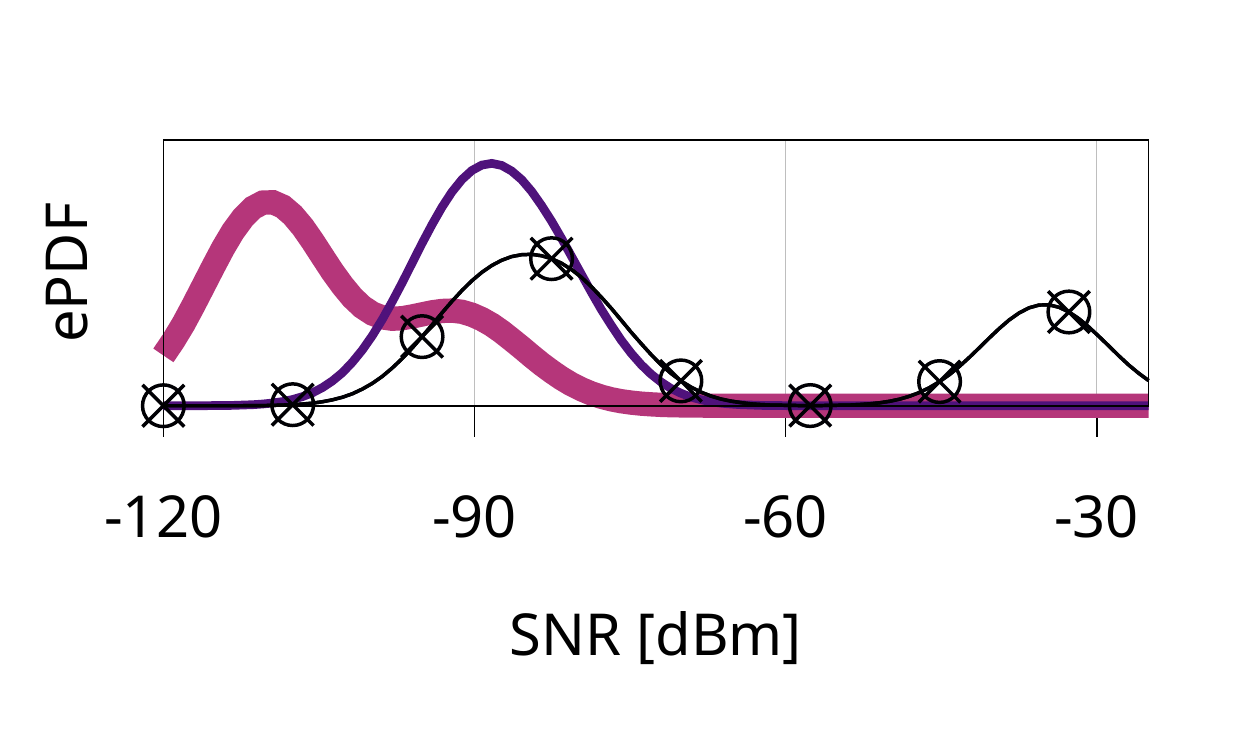} }}~
    \subfloat[\label{fig:compare-delay} ]{{\includegraphics[trim=13 21 18 10, clip,width=.42\columnwidth]{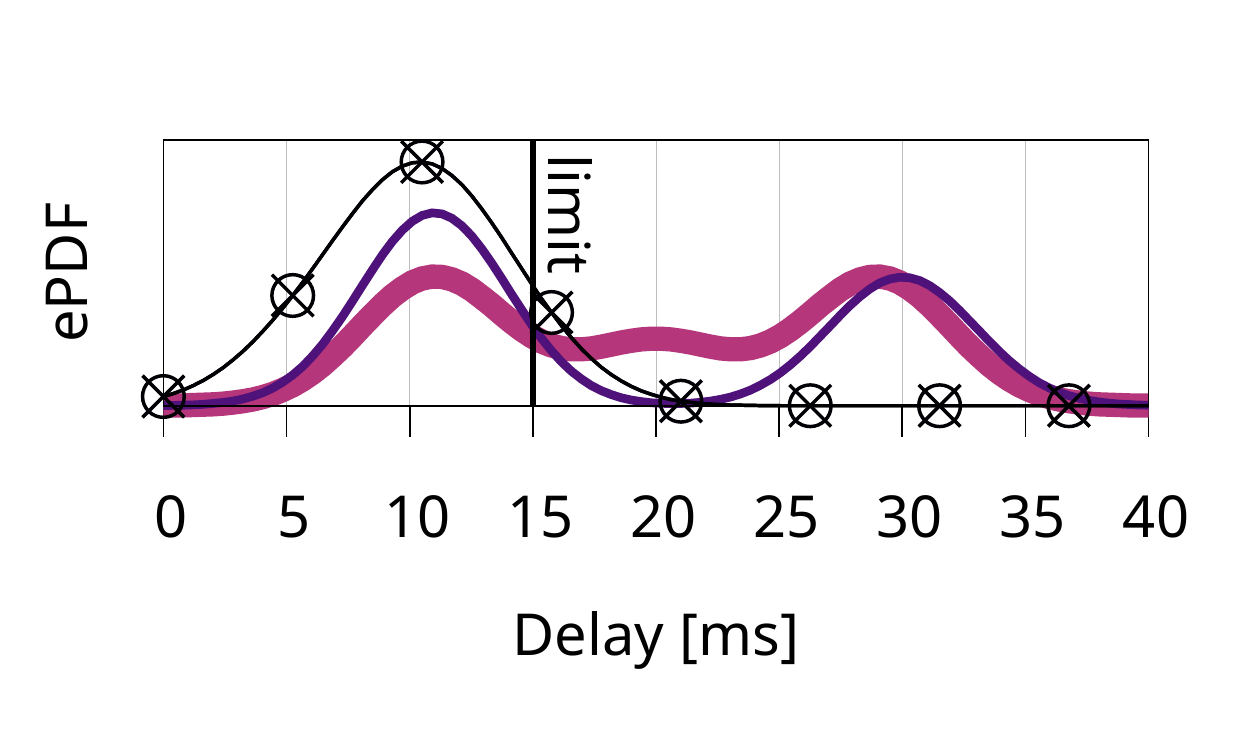} }}\\\vspace{-.7em}
    \subfloat[\label{fig:compare-cost}  ]{{\includegraphics[trim=13 21 18 10, clip,width=.42\columnwidth]{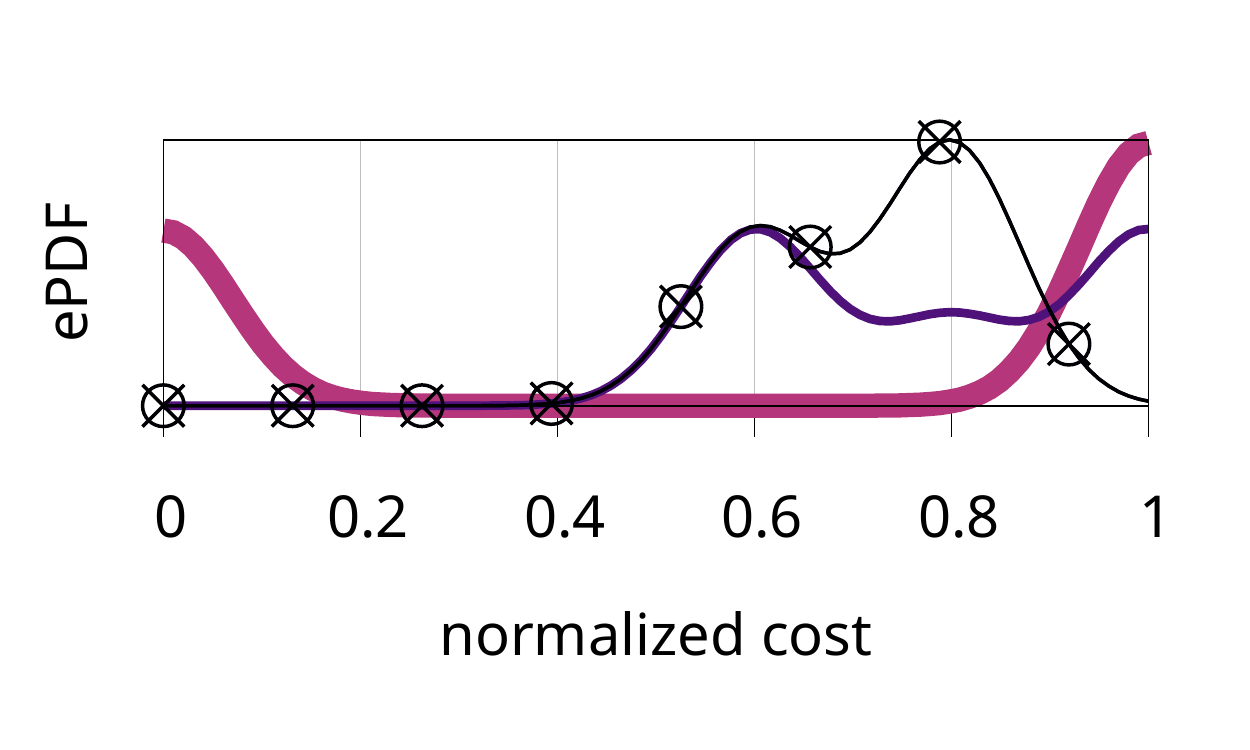} }}~
    \subfloat[\label{fig:compare-bw}    ]{{\includegraphics[trim=13 21 18 10, clip,width=.42\columnwidth]{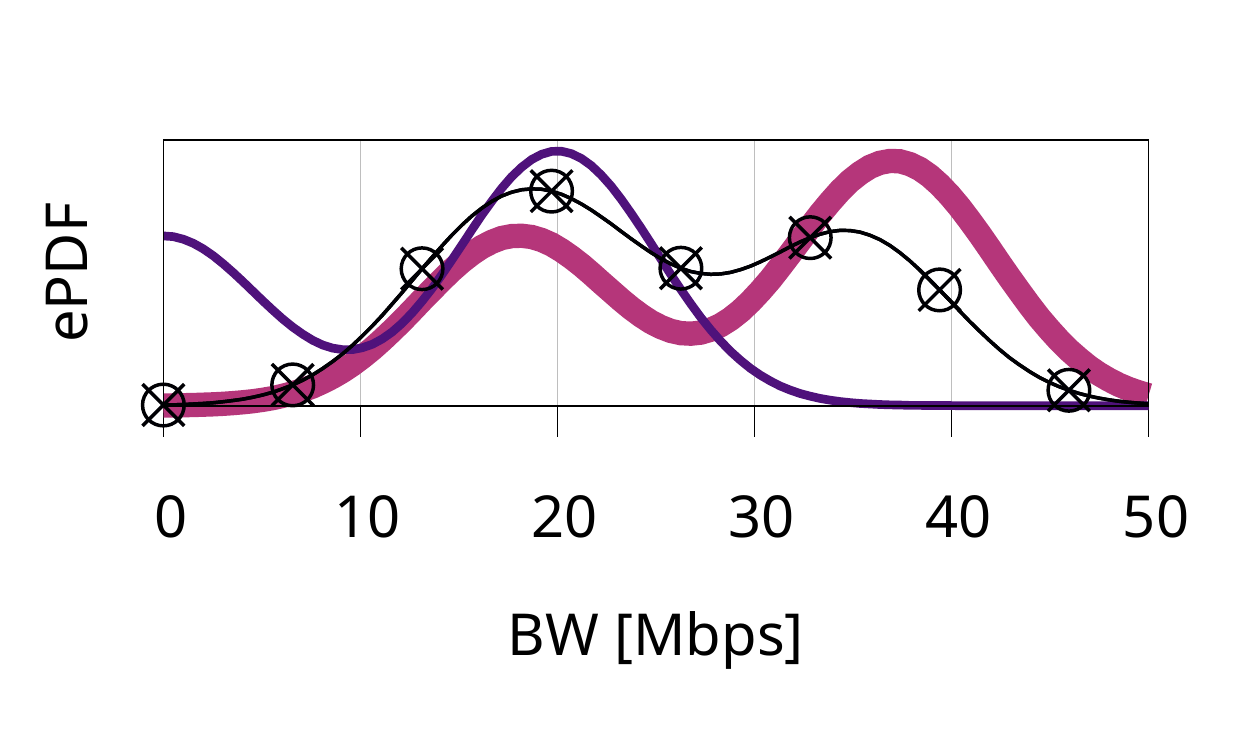} }}~

   \caption{ePDFs for the experienced
    SNR, Delay, cost, and bandwidth
    consumption during Fig.~\ref{fig:corridor}
    driving.
    Plots illustrate the experienced metrics
    using DLMD (circle),
    optimal (cross),
    \cite{delgado2022oros} (thickest), and
    \cite{delayandreliability} (thick)
   }%
   \label{fig:compare}%
\end{figure}

For the sake of comparison, in Fig.~\ref{fig:compare}
we report the ePDF of different metrics considered in the problem.
Overall, results show that DLMD
attains better SNR than 
\cite{delgado2022oros,delayandreliability}
and remains below the 15~ms
delay limit. Moreover, DLMD matches the
metrics achieved by the optimal solution
-- see how circle and cross markers overlap
in Fig.~\ref{fig:compare}.
In terms of cost
and bandwidth consumption, DLMD
achieves a nice trade-off with respect to other
solutions because it only migrates resources to
the expensive Edge when needed.

\subsection{Stress Tests}

\begin{figure}[t]
    \subfloat[\label{fig:pareto} ]{{\includegraphics[trim=16 11 21 23,clip,width=0.5\columnwidth]{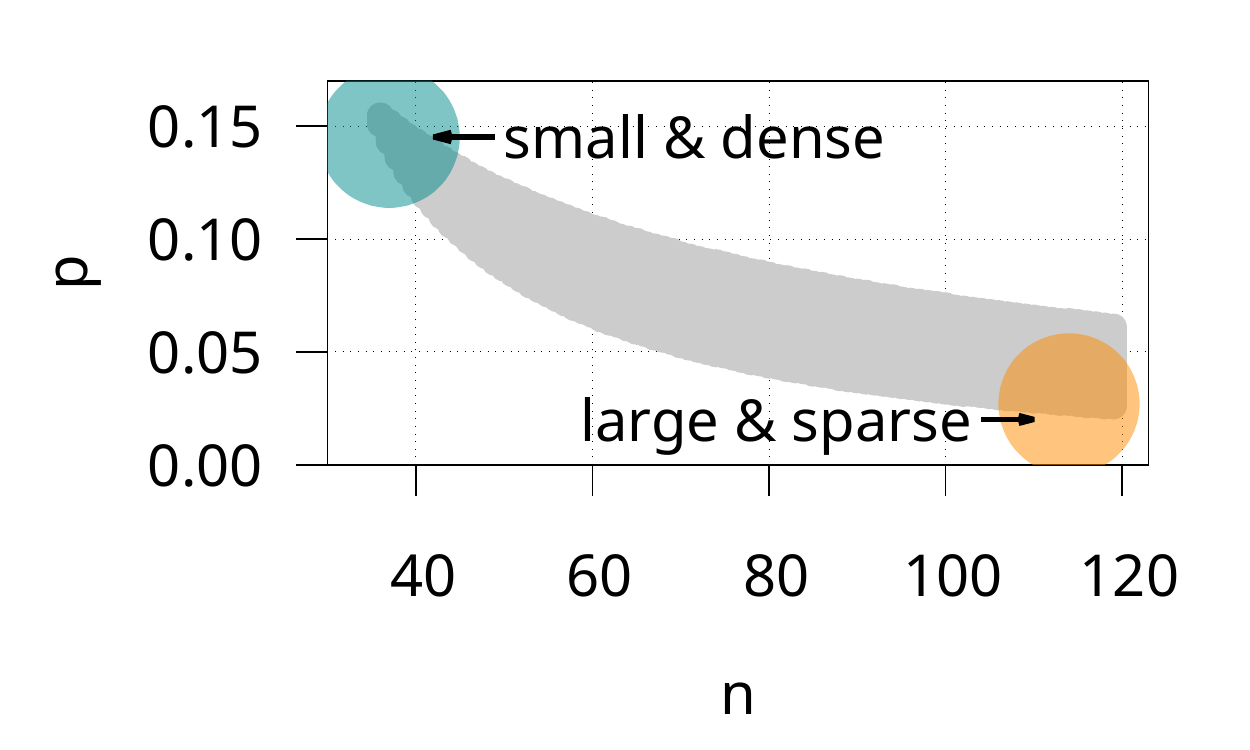} }}~\hspace{-2em}~
    \subfloat[\label{fig:small-dense} ]{{%
        \includegraphics[trim=150 1 150 1,clip,width=0.11\columnwidth]{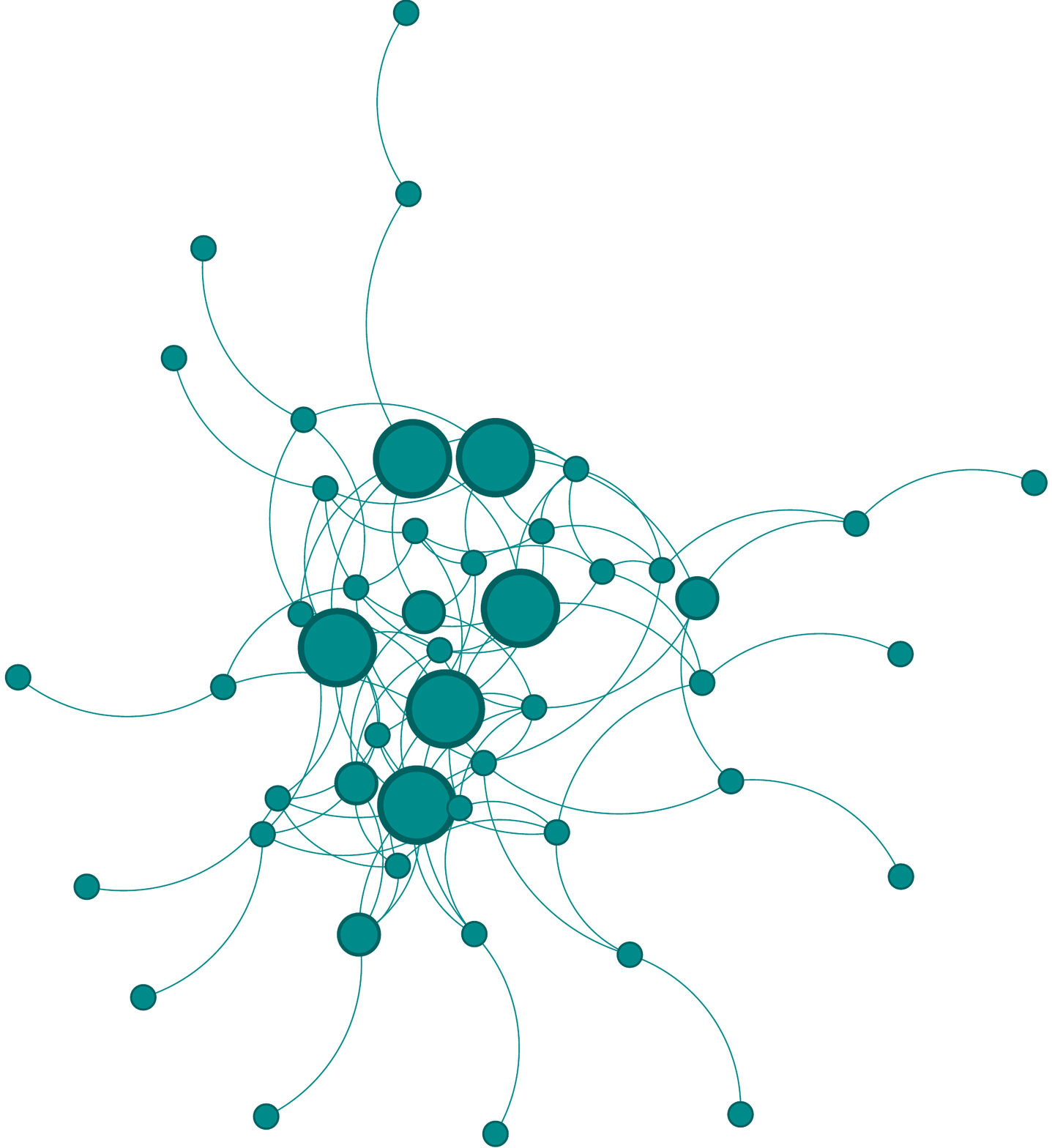}\\
    \hspace{3em}\includegraphics[trim=125 1 150 1,clip,width=0.17\columnwidth]{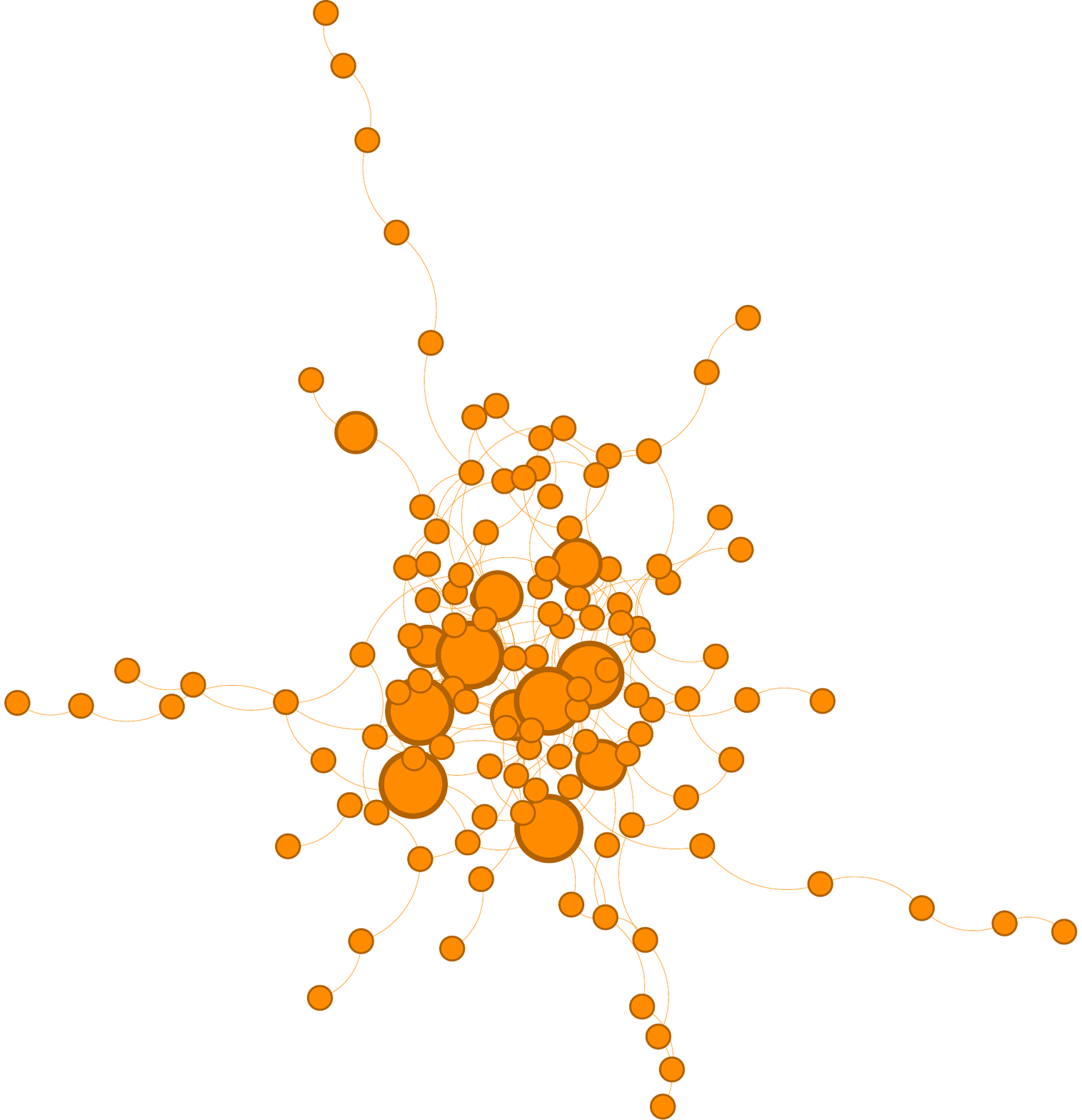} }}~\hspace{-3em}~
    \subfloat[\label{fig:urtinsa} ]{{\includegraphics[width=0.17\columnwidth]{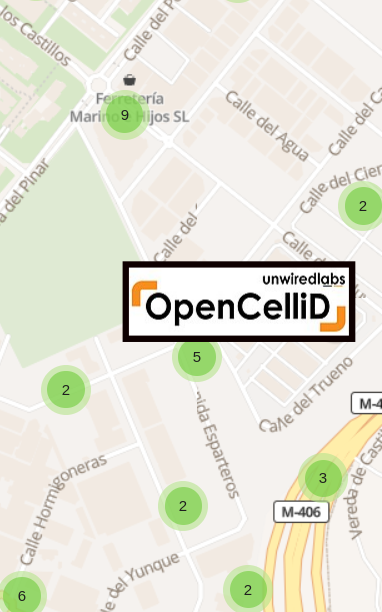} }}
    
    \caption{Erdős–Rényi setups
    (a) for realistic network
    graphs (b) that conect OpenCellid
    PoAs of an
    industrial area (c).}
    \label{fig:industrial-setup}
\end{figure}

In this section we assess DLMD performance upon scarce
of network resources, i.e., when the network
is stressed. To do so, we collect the PoAs present
in an industrial area of Alcorcón, Spain; and
generate a small \& large random graph that conveys
the network topology
-- see~Fig.~\ref{fig:industrial-setup}.

Specifically, we derive $V(G)$ using
Erdős–Rényi $G(n,p)$ graphs with $n=48,128$ nodes;
and find adequate $p$ to have:
6 Cloud servers with 6;
4 far Edge servers with 4; and
2 near Edge servers with 2 redundant links.
With this information, it is possible to find
feasible $(n,p)$ setups -- see~Fig.~\ref{fig:pareto}
gray region -- knowing that
$\mathbb{P}(deg(v)= k)= {n-1 \choose p}p^k(1-p)^{n-1-k}$.

\begin{figure}[t]
    \centering
    \subfloat[\label{fig:stress-delay} ]{{\includegraphics[trim=13 27 14 10, clip,width=.44\columnwidth]{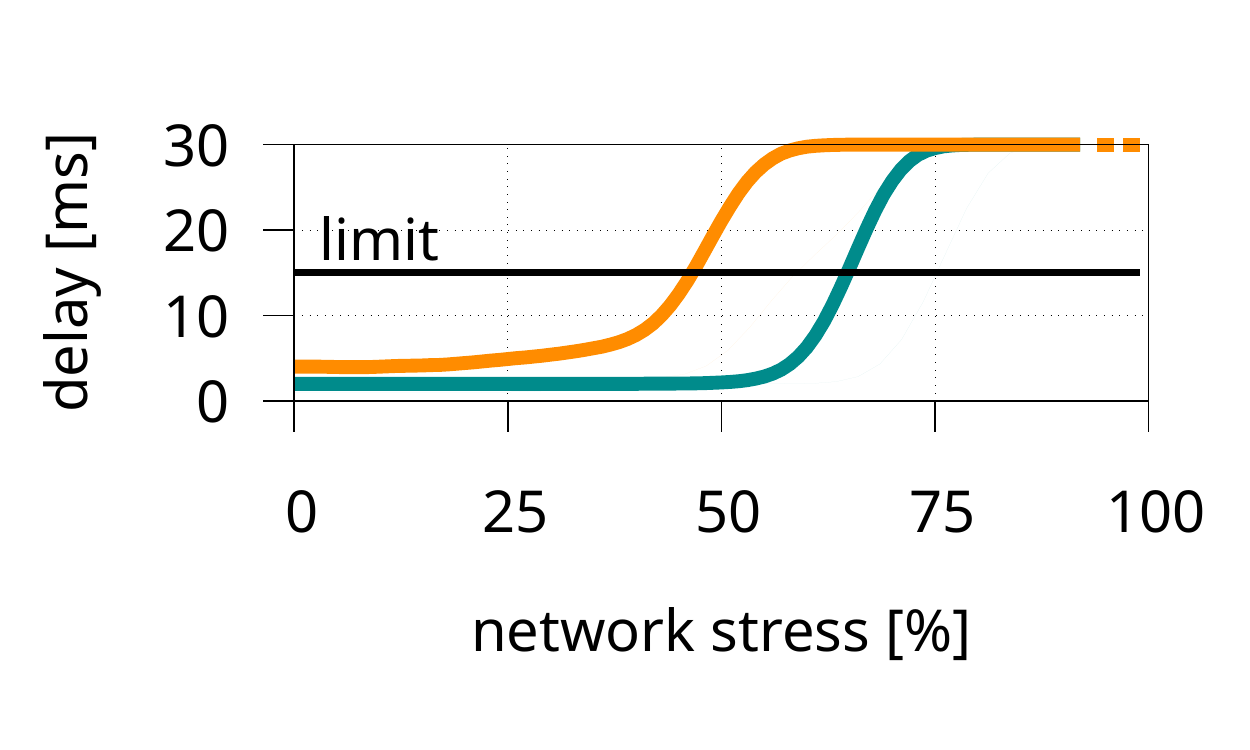} }}~
    \subfloat[\label{fig:stress-edge} ]{{\includegraphics[trim=13 27 14 10, clip,width=.44\columnwidth]{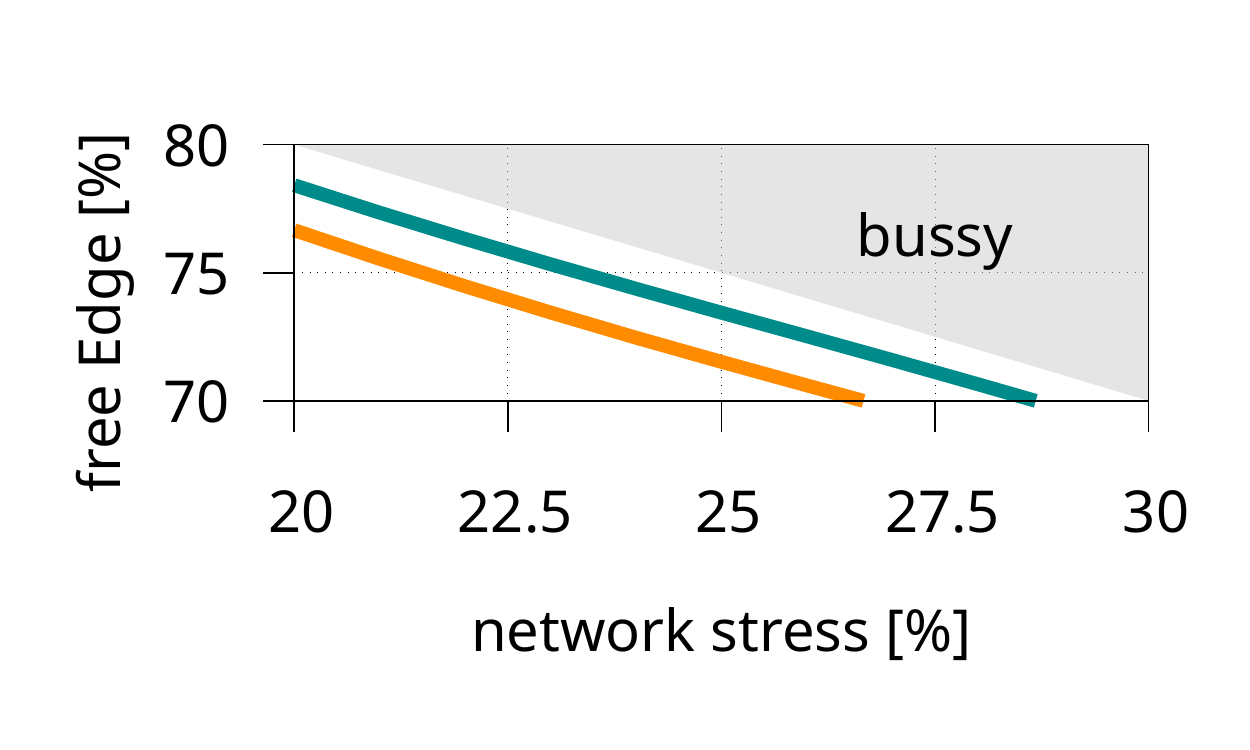} }}\\
    \subfloat[\label{fig:stress-migrations} ]{{\includegraphics[trim=13 27 14 10, clip,width=.44\columnwidth]{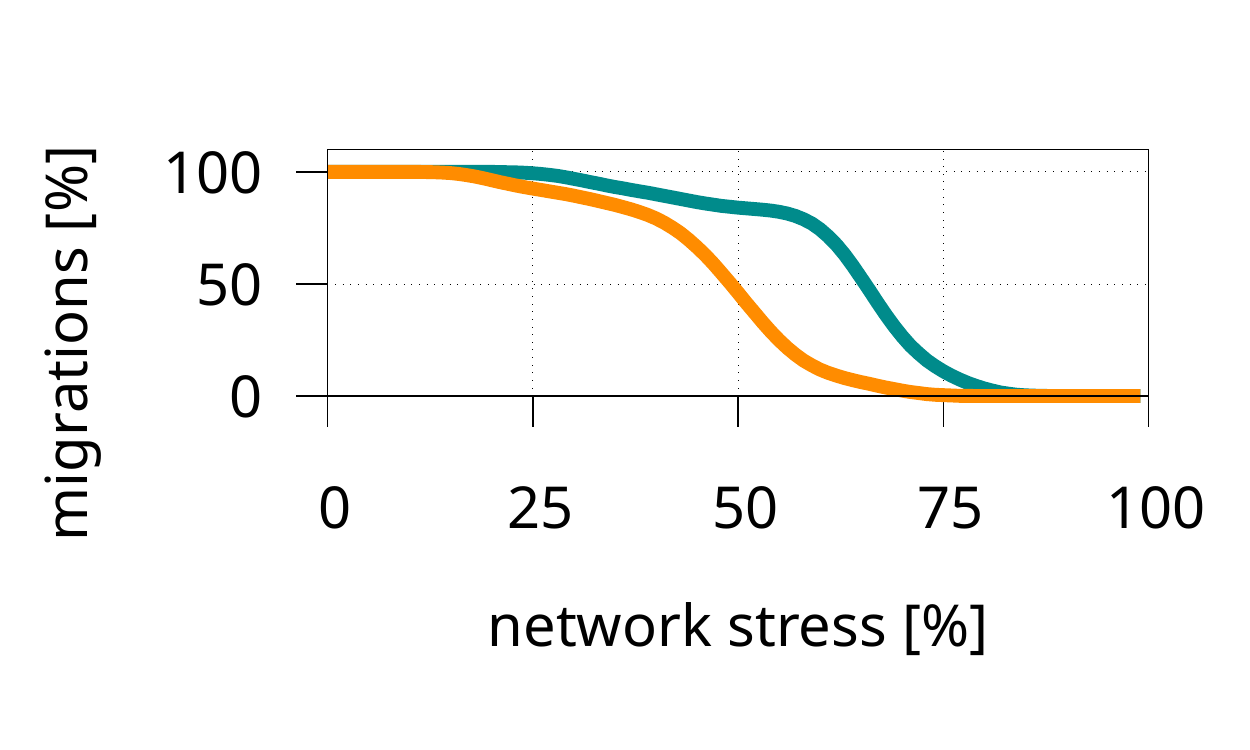} }}~
    \subfloat[\label{fig:stress-runtime} ]{{\includegraphics[trim=13 27 14 10, clip,width=.44\columnwidth]{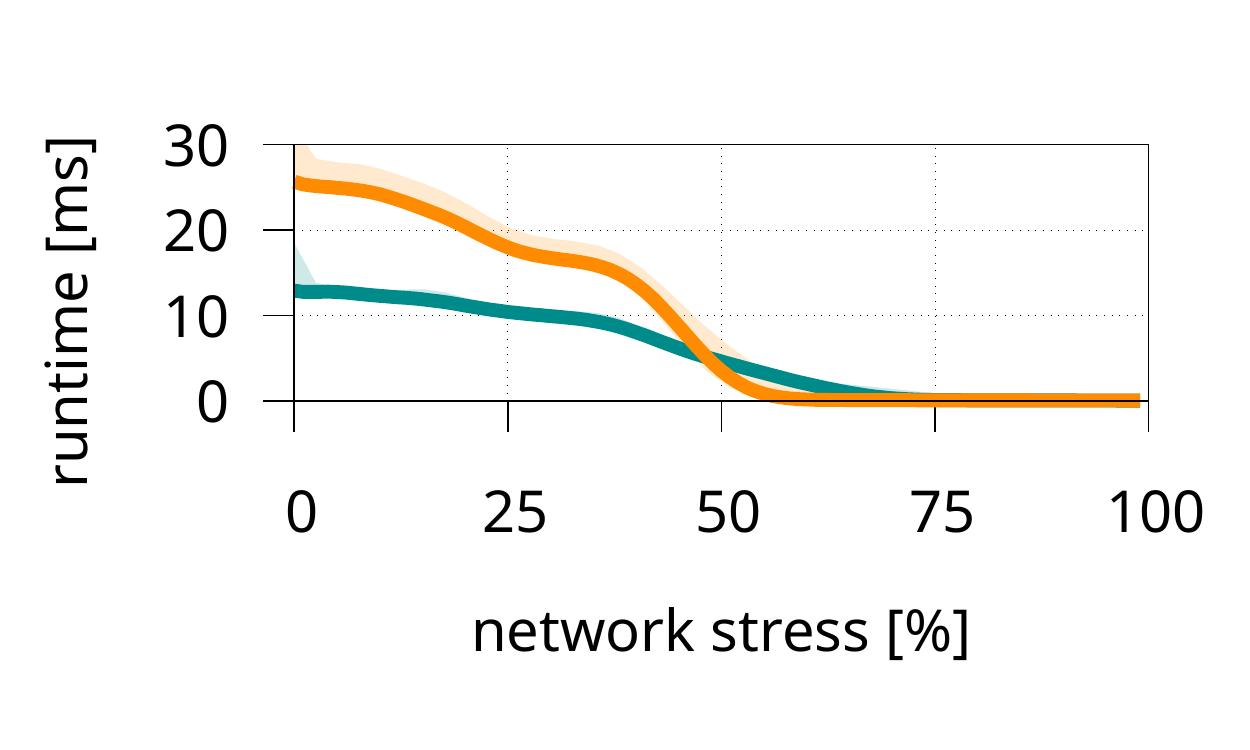} }}~

   \caption{DLMD stress test
    in Fig.~\ref{fig:industrial-setup}
    small (green) and large (orange)
    networks.
    Tests show 90\% confidence intervals
    (shade).}%
   \label{fig:stress}%
\end{figure}

Fig.~\ref{fig:stress} shows how DLMD behaves as
the network usage/stress evenly increases from 0
to a 100\% in small
\& dense graphs (green), and large \& sparse
graphs (orange). In the experiment we use an
enriched robot service with 3 VFs being offloaded
to remote servers. As the robot moves along the
12~PoAs taken from OpenCelliD, DLMD performs
migrations and handover decisions to keep an
adequate connectivity between the robot and the
offloaded VFs.

Fig.~\ref{fig:stress-delay} evidences that DLMD meets the delay service requirement of 15ms as long as network stress is below 40\%.
In smaller and dense graphs, the delay
requirement is satisfied even with 60\%
stress as it has higher migration
success thanks to the graph density
-- see Fig.~\ref{fig:stress-migrations}.

Lastly, it is worth mentioning that during the
experiments DLMD uses less than a 5\% of the
available Edge resources, for it exploits the
Cloud as long as it is close enough and available
-- see Fig.~\ref{fig:stress-edge}.
Moreover, DLMD managed to find solutions in less than
30~ms in both small and large graphs
-- see Fig.~\ref{fig:stress-runtime}.

\section{Conclusion}
\label{sec:conclusion}
In this paper we present DLMD, an
offloading algorithm for networked
robotics that fosters offloading
VFs to the Cloud to minimize the Edge
usage. Additionally, DLMD
assists in the VF migration
and radio handover as the robot moves.
Results demonstrate that DLMD $(i)$ outperforms state of the art solutions in small warehousing scenarios; $(ii)$ takes $\leq30$ms to find solutions in small and large graphs with real-world PoA locations; and $(iii)$ achieves
the goal of minimizing resource consumption
at the Edge despite the network stress.
To the best of our knowledge, this is the first work to consider latency, radio signal, and robot mobility simultaneously to tackle the VF embedding problem.
Future work will focus on larger
scenarios and a weighted
$\tfrac{1}{\lambda}+d$ metric for DLMD.


\bibliographystyle{IEEEtran}
\bibliography{main}

\end{document}